\documentclass[runningheads]{llncs}
\usepackage[width=122mm,left=12mm,paperwidth=146mm,height=193mm,top=12mm,paperheight=217mm]{geometry}
\usepackage{tikz}

\usepackage{graphicx}
\usepackage{comment}
\usepackage{amsmath,amssymb} 
\usepackage{mathrsfs}
\usepackage{bm}
\usepackage{algorithm}
\usepackage{algpseudocode}
\usepackage{dsfont}
\usepackage{color}
\usepackage{multirow}
\usepackage{cite}
\usepackage[pagebackref=true,breaklinks=true,letterpaper=true,colorlinks,bookmarks=false]{hyperref}
\hypersetup{
    colorlinks = true,
    citecolor  = blue,
    linkcolor  = blue,
    urlcolor   = blue,
}
\usepackage{cleveref}
\usepackage{courier}
\Crefformat{figure}{Fig.~#2#1#3}                           
\Crefname{subfigure}{Fig.}{Figs.}
\Crefname{figure}{Fig.}{Figs.}
\usepackage{booktabs} 

\usepackage{subfig}
\usepackage{caption}
\graphicspath{{./figure/}}
\usepackage{filecontents}                                  
\usepackage{pgfplots}
\usepackage{pgfplotstable}
\pgfplotsset{compat=newest}

\DeclareRobustCommand\onedot{\futurelet\@let@token\@onedot}
\def\@onedot{\ifx\@let@token.\else.\null\fi\xspace}

\begin{document}
\pagestyle{headings}
\mainmatter
\def\ECCVSubNumber{4095}  

\title{
    Dive Deeper Into Box for Object Detection
}

\titlerunning{Dive Deeper Into Box for Object Detection}
%
\author{Ran Chen\inst{1} \and
Yong Liu\inst{2} \and
Mengdan Zhang\inst{2} \and
Shu Liu\inst{3} \and \\
Bei Yu \inst{1}\and
Yu-Wing Tai\inst{4} }
\authorrunning{R. Chen et al.}
%
\institute{
    The Chinese University of Hong Kong\\
    \and
    Tencent Youtu Lab\\
    \and
    SmartMore\\
    \and
    The Hong Kong University of Science and Technology\\
}
\maketitle

\begin{abstract}

Anchor free methods have defined the new frontier in state-of-the-art object detection researches where accurate bounding box estimation is the key to the success of these methods. However, even the bounding box has the highest confidence score, it is still far from perfect at localization. 
To this end, we propose a box reorganization method (DDBNet), which can dive deeper into the box for  more accurate localization.
At the first step, drifted boxes are filtered out because the contents in these boxes are inconsistent with target semantics. Next, the selected boxes are broken into boundaries, and the well-aligned boundaries are searched and grouped into a sort of optimal boxes toward tightening instances more precisely.  
Experimental results show that our method is effective which leads to state-of-the-art performance for object detection.
\end{abstract}

\section{Introduction}

Object detection is an important task in computer vision, which requires predicting a bounding box of an object with a category label for each instance in an image. 
State-of-the-art techniques can be divided into either anchor-based methods \cite{OD-CVPR2014-RCNN,OD-ICCV2015-FastRCNN,OD-TPAMI2017-FasterRCNN,OD-arXiv2017-DSSD,OD-ECCV2016-Liu,OD-CVPR2018-Cai,he2017mask,OD-CVPR2017-YOLO9000,redmon2018yolov3}
and anchor-free methods \cite{OD-CVPR2016-YOLO, OD-ICCV2019-FCOS, OD-ICCV2019-CenterNet, OD-CVPR2019-Zhou, OD-ICCV2019-RepPoints, jie2016scale}. 
Recently, the anchor-free methods have increasing popularity over the anchor-based methods in many applications and benchmarks\cite{lin2014microsoft, everingham2015pascal, deng2009imagenet, OD-CVPR2012-Geiger}.
Despite the success of anchor-free methods, 
one should note that these methods still have limitations on their accuracy, which are bounded by the way that the bounding boxes are learned in an atomic fashion. 
Here, we discuss two concerns of existing anchor-free methods which lead to the inaccurate detection.

\begin{figure}[tb!]
    \centering
    \includegraphics[width=0.7\linewidth]{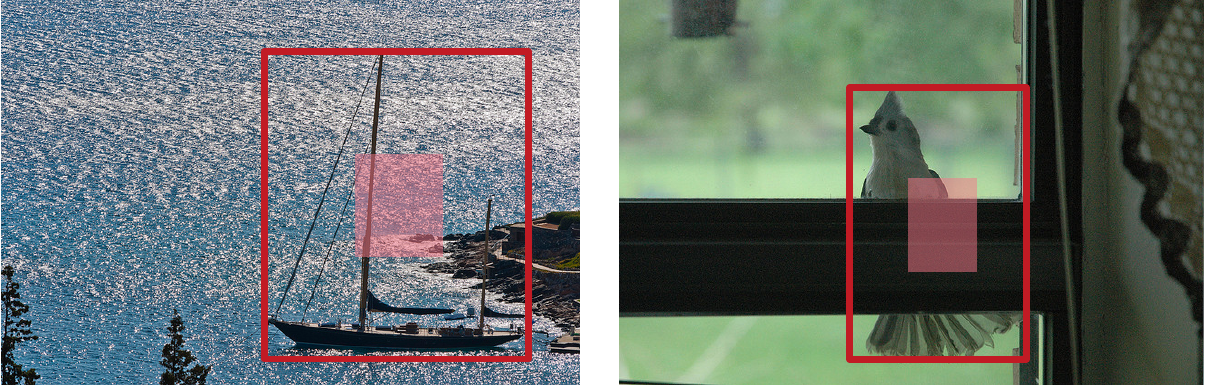} 
    \caption{\textbf{An illustration of the inconsistency between the semantics of center key-points inside a bounding box and their annotations.} Pixels of backgrounds in the red central area are considered as positive center key-points, which is incorrect.}
    \label{figure:inconsistency-example}
\end{figure}

    First, the definition of center key-points \cite{OD-ICCV2019-CenterNet} is inconsistent with their semantics.
    As we all know that center key-point is essential for anchor-free detectors. It is a common strategy to embed positive center key-points inside an object bounding box into a Uniform or Gaussian distribution in the training stage of the anchor-free detectors such as FCOS \cite{OD-ICCV2019-FCOS} and CornerNet \cite{OD-ECCV2018-CornerNet}. 
    However, it is inevitable to falsely consider noisy pixels from background as positives, as illustrated in \Cref{figure:inconsistency-example}.
    Namely, exploiting a trivial strategy to define positive targets would lead to a significant semantic inconsistency, degrading the regression accuracy of detectors.  
    
\begin{figure}[tb!]
    \centering
    \includegraphics[width=0.66\linewidth]{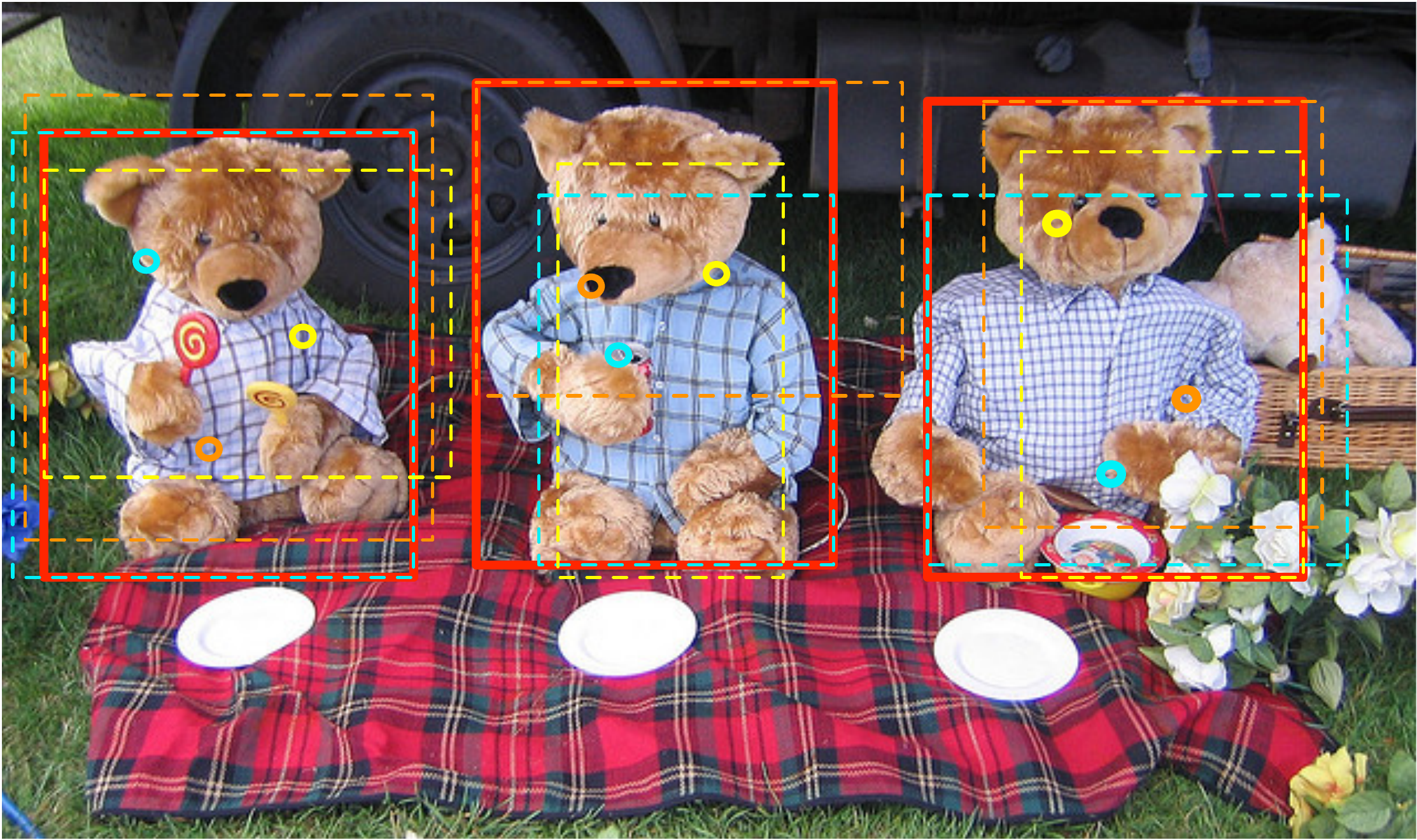} 
    \caption{\textbf{An illustration of the boundary drifts in box predictions of general anchor-free detectors.} Limited by regional receptive fields and the design of treating each box prediction as an atomic operation in general detectors, each predicted box with dotted line is imperfect where four boundaries are not well aligned to the ground truth simultaneously. After box decomposition and combination, the reorganized box with red color gets better localization.}
    \label{figure:D_R-visualize}
\end{figure}

    
    Second, local wise regression is limited. Concretely, a center key-point usually provides box predictions in a regional/local-wise manner, which potentially defects the detection accuracy. The local-wise prediction results from the limitation of the receptive fields of convolution kernels, and the design of treating each box prediction from each center key-point as an atomic operation. 
    As shown in \Cref{figure:D_R-visualize}, the dotted predicted box and corresponding center key-point are presented in the same color.
    Although each predicted box is surrounding the object, it is imperfect because four boundaries are not well aligned to the ground truth simultaneously.
    As a result, choosing a box of high score at inference stage as the final detection result is sometimes inferior.

To tackle the inaccurate detection problem, we present a novel bounding box reorganization method, which dives deeper into box regressions of center key-points and takes care of semantic consistencies of center key-points.
This reorganization method contains two modules, denoted as box decomposition and recombination (D\&R) module and semantic consistency module.
Specifically, box predictions of center key-points inside an instance form an initial coarse distribution of the instance localization. This distribution is not well aligned to the ideal instance localization, and boundary drifts usually occur. 
The D\&R module is proposed to firstly decompose these box predictions into four sets of boundaries to model an instance localization  at a lower refined level, where the confidence of each boundary is evaluated according to the deviation with ground-truth. 
Next, these boundaries are sorted and recombined to form a sort of more accurate box predictions for each instance, as described in \Cref{figure:D_R-visualize}.
Then, these refined box predictions contribute to the final evaluation of box regressions.

Meanwhile, the semantic consistency module is proposed to rule out noisy center key-points coming from the background, which allows our method to focus on key-points that are strongly related to the target instance semantically.
Thus, box predictions from these semantic consistent key-points can form a more tight and robust distribution of the instance localization, which further boosts the performance of the D\&R module.
Our semantic consistency module is an adaptive strategy without extra hyper-parameters for predefined spatial constraints, which is superior to existing predefined strategies in \cite{OD-CVPR2019-Zhucc, OD-CVPR2019-Wang, OD-ICCV2019-FCOS}.

The main contribution of this work lies in the following aspects.
\begin{itemize}
    \item We propose a novel box reorganization method in a unified anchor free detection framework. Especially, a D\&R module is proposed to
    take the boundary prediction as an atomic operation, and then reorganize well-aligned boundaries into boxes in a bottom-up fashion with negligible computation overhead.
    To the best of our knowledge, the idea of breaking boxes into boundaries for training has never been investigated in this task.
    \item We evaluate the semantic inconsistency between center key-points inside an instance and the annotated labels, which helps boost  the convergence of a detection network.
    \item The proposed method DDBNet obtains a state-of-the-art result of 45.5\% in AP. The stable experimental results in all metrics ensure that this method can be effectively extended to typical anchor free detectors.
\end{itemize}


\section{Related Work}

\begin{figure*}[tb!]
    \centering
     \includegraphics[width=0.95\linewidth]{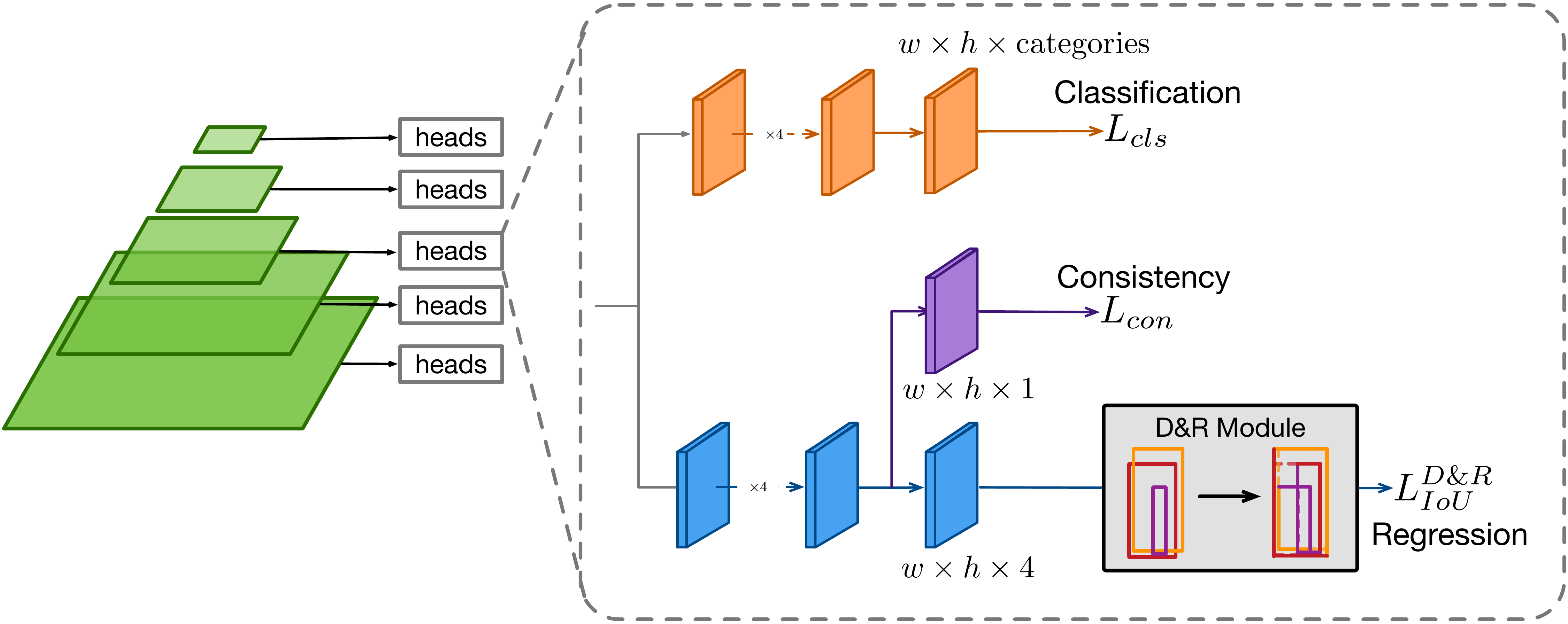} 
    \caption{\textbf{An illustration of our network architecture.} Two novel components: the D\&R module and the consistency module are incorporated into a general detection network. The D\&R module carries out box decomposition and recombination in the training stage regularized by the IoU loss and predicts boundary confidences supervised by the boundary deviation. The consistency module selects meaningful pixels whose semantics is consistent with the instance to improve network convergence in the training stage.
    }
    \label{figure: network-structure}
\end{figure*} 

\noindent\textbf{Anchor based Object Detectors.}
In anchor-based detectors, the anchor boxes can be viewed as pre-defined sliding windows or proposals, which are classified as positive or negative samples, with an extra offsets regression to refine the prediction of bounding boxes.
The design of anchor boxes is popularized by two-stage approaches such as Faster R-CNN in its RPNs \cite{OD-TPAMI2017-FasterRCNN}, and single-stage approaches such as SSD \cite{OD-ECCV2016-Liu}, RetinaNet \cite{OD-ICCV2017-RetinaNet}, and YOLO9000 \cite{OD-CVPR2017-YOLO9000}, which has become the convention in a modern detector.
Anchor boxes make the best use of the feature maps of CNNs and avoid repeated feature computation, speeding up the detection process dramatically.
However, anchor boxes result in excessively too many hyper-parameters that are used to describe anchor shapes or to label each anchor box as a positive, ignored or negative sample.
These hyper-parameters have shown a great impact on the final accuracy, and require heuristic tuning.


\noindent\textbf{Anchor Free Object Detectors.}
Anchor-free detectors directly learn the object existing possibility and the bounding box coordinates without anchor reference.
DenseBox \cite{OD-arXiv2015-DenseBox} is a pioneer work of anchor-free based detectors. While due to the difficulty of handling overlapping situations, it is not suitable for generic object detection.

One successful family of anchor free works \cite{OD-ICCV2019-FCOS,OD-CVPR2019-Zhucc,OD-arXiv2019-FoveaBox,OD-CVPR2019-Wang} adopts the Feature Pyramid network \cite{OD-CVPR2017-FPN} (FPN) as the backbone network and applies direct regression and classification on multi-scale features. These methods treat the bounding box prediction as an atomic task without any further investigations, which bounds the detection accuracy due to the two concerns we discussed in the introduction. 
To avoid the drawback of anchors and refine the box presentations, points based box representation becomes popular recently \cite{OD-ECCV2018-CornerNet,OD-CVPR2019-Zhou,OD-ICCV2019-RepPoints,OD-ICCV2019-CenterNet}. For example, CornerNet \cite{OD-ECCV2018-CornerNet} predicts the heatmap of corners and apply an embedding method to group a pair of corners that belong to the same object. \cite{OD-CVPR2019-Zhou} presents a bottom-up detection framework inspired by the keypoint estimations. Compared to these points based methods, our proposed method has following innovations: 1) Our method focuses on the mid-level boundary representations to achieve a balance between accuracy and robustness of feature modeling; 2) Our method does not need to learn an embedding explicitly while obtaining a reliable boundary grouping to produce the final bounding box predictions.

Furthermore, it is observed that anchor-free methods may produce a number of low-quality predicted bounding boxes at locations that are far from the center of a target object.
In order to suppress these low-quality detections, a novel ``centerness'' branch to predict the deviation of a pixel to the center of its corresponding bounding box is exploited in FCOS \cite{OD-ICCV2019-FCOS}.
This score is then used to down-weight low-quality detected bounding boxes and merge the detection results in NMS. FoveaBox \cite{OD-arXiv2019-FoveaBox} focuses on the object's center motivated by the fovea of human eyes.
It is reasonable to degrade the importance of pixels close to boundaries, but the predefined center field may not cover all cases in the real world, as shown in \Cref{figure:inconsistency-example}.
Thus, we propose an adaptive consistency module to solve the inconsistency issue mentioned above between the semantics of pixels inside an instance and the predefined labels or scores.

\section{Our Approach}
In this work, we build DDBNet based on FCOS as a demonstration, which is an advanced anchor-free method. As shown in \Cref{figure: network-structure}, our innovations lie in the box decomposition and recombination (D\&R) module and the semantic consistency module. 

To be specific, the D\&R module reorganizes the predicted boxes by breaking them into boundaries for training which is concatenated behind the regression branch. In the training stage, once bounding box predictions are regressed at each pixel, the D\&R module decomposes each bounding box into four directional boundaries. Then, boundaries of the same kind are ranked by their actual boundary deviations from the ground-truth. Consequently, by recombining ranked boundaries, more accurate box predictions are expected, which are then optimized by the IoU loss \cite{yu2016unitbox}.

As for the semantic consistency module, a new branch of estimating semantic consistency instead of centerness is incorporated into the framework, which is optimized under the supervision of the semantic consistency module. This module exploits an adaptive filtering strategy based on the outputs of the classification and the regression branches. More details about the two modules are provided in the following subsections.

\subsection{Box Decomposition and Recombination}
\label{subsec:box-decom-recom}

Given an instance $I$, every pixel $i$ inside of $I$ regresses a box $p_{i} = \{l_{i}, t_{i}, r_{i}, b_{i}\}$.
The set of predicted boxes is denoted as $B_{I} = \{ p_{0}, p_{1}, \dots, p_{n}\}$, where $l, t, r, b$ denote the left, the top, the right, and the bottom boundaries respectively.


\begin{figure*}[tb!]
    \centering
    \includegraphics[width=1.0\linewidth]{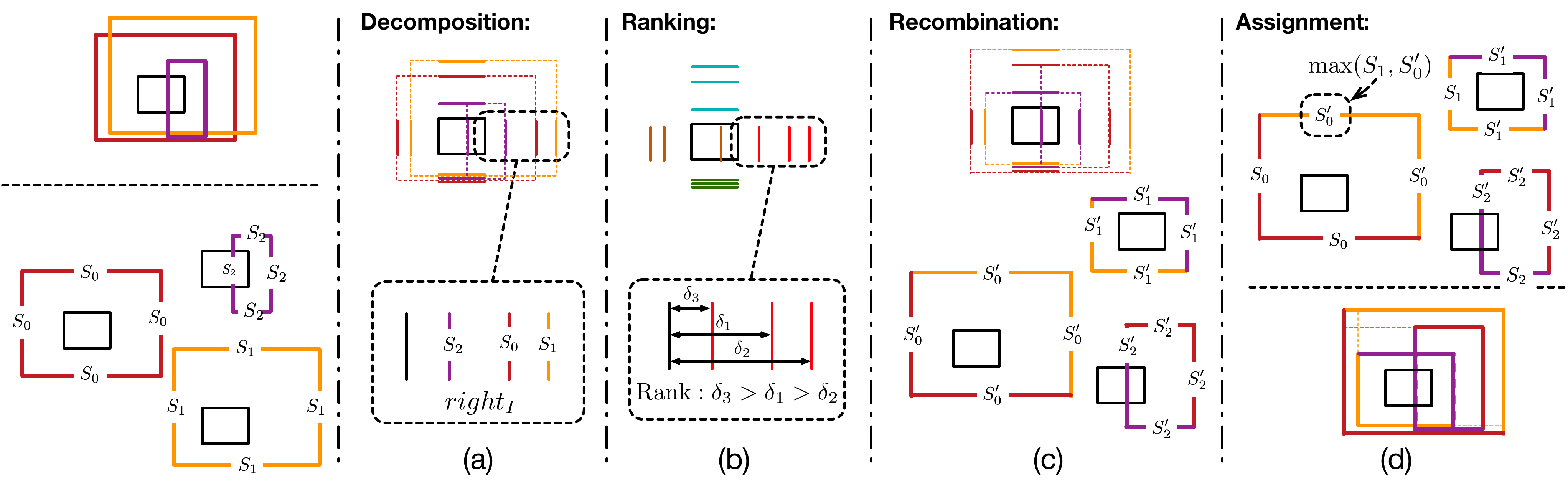} 
    \caption{\textbf{An illustration of the work flow of the D\&R module.}
    For a clear visualization, only three predictions in color are provided for the same ground-truth shown in black.
    (a) \textbf{Decomposition:} Break up boxes and assign IoU scores $S_{0}, S_{1}, S_{2}$ of boxes to boundaries as confidence.
    (b) \textbf{Ranking:} The rule how we recombine boundaries to new boxes.
    (c) \textbf{Recombination:} Regroup boundaries as new boxes and assign new IoU scores $S'_{0}, S'_{1}, S'_{2}$ to boundaries as confidence.
    (d) \textbf{Assignment:} Choose the winner confidence as final result.
    The recombined box is shown on the right.}
    \label{figure:D_R-flow}
\end{figure*} 

Normally, an IoU regression loss is expressed as 
\begin{equation}
    \label{loss: origin-iou}
    L_{IoU} = -\frac{1}{N_{pos}}\sum_{I}\sum_{i}^{n} \log(IoU(p_{i}, p^{*}_{I})),
\end{equation} 
where $N_{pos}$ is the number of positive pixels of all instances, $p^{*}_{I}$ is the regression target. Simply, the proposed box decomposition and recombination (D\&R) module is designed to reproduce more accurate $p_{i}$ with the optimization of IoU loss. As shown in \Cref{figure:D_R-flow}, the D\&R module consists of four steps before regularizing the final box predictions based on the IoU regression. More details are described as follows.

\noindent\textbf{Decomposition:} 
A predicted box $p_{i}$ is split into boundaries $l_{i},t_{i},r_{i},b_{i}$ and the IoU $s_{i}$ between $p_{i}$ and $p^{*}_{I}$ is assigned as the confidences of four boundaries, as shown in \Cref{figure:D_R-flow}(a).
For instance $I$, the confidences of boundaries is denoted as a $N \times 4$ matrix $S_{I}$.
Then we group four kinds of boundaries into four sets, which are $left_{I} = \{ l_{0}, l_{1}, ..., l_{n}\}$, $right_{I} = \{ r_{0}, r_{1}, ..., r_{n}\}$, $bottom_{I} = \{ b_{0}, b_{1}, ..., b_{n}\}$, $top_{I} = \{ t_{0}, t_{1}, ..., t_{n}\}$. 

\noindent\textbf{Ranking:} 
Considering the constraint of the IoU loss \cite{yu2016unitbox}, where the larger intersection area of prediction boxes with smaller union area is favored, the optimal box prediction is expected to have the lowest IoU loss. Thus, traversing all the boundaries of the instance $I$ to obtain the optimal box rearrangement $B'_{I}$ is an intuitive choice. However, in this way, the computation complexity is quite expensive, which is $\mathcal{O}(n^4)$. To avoid the heavy computation brought by such brute force method, we apply a simple and efficient ranking strategy.
For each boundary set of instance $I$, the deviations $\delta_{I}^{_l}$, $\delta_{I}^{_r}$, $\delta_{I}^{_b}$, $\delta_{I}^{_t}$ to the targets boundary $p^{*}_{I} = \{ l_{I}, r_{I}, b_{I}, t_{I}\}$ are calculated. Then, boundaries in each set are sorted by the corresponding deviations, as shown in \Cref{figure:D_R-flow}(b). The boundary closer to the ground-truth has the higher rank than the boundary farther. We find that this ranking strategy works well and the ranking noise does not affect the stability of the network training. 

\noindent\textbf{Recombination:} 
As shown in \Cref{figure:D_R-flow}(c), boundaries of four sets with the same rank are recombined as a new box $B'_I = \{ p'_{0}, p'_{1}, \dots, p'_{n}\}$.
Then the IoU $s'_{i}$ between $p'_{i}$ and $p^{*}_{I}$ is assigned as the recombination confidence of four boundaries. 
The confidences of recombination boundaries is expressed as matrix $S'_{I}$ with shape $N \times 4$.

\noindent\textbf{Assignment:} 
Now we get two sets of boundaries scores $S_I$ and $S'_{I}$. As described as \Cref{figure:D_R-flow}(d), the final confidence of each boundary is assigned using the higher score within $S_I$ and $S'_{I}$ instead of totally using $S'_{I}$. This assignment strategy results from the following case, \textit{e.g}.~the recombined low-rank box contains boundaries far away from the ground-truth. Then, the confidences $s'_{i}$ of four boundaries after recombination are much lower than their original one $s_{i}$. The severely drifted confidence scores lead to unstable gradient back-propagation in the training stage. 

Thus, for reliable network training, each boundary is optimized under the supervision of the IoU loss estimated based on the ground-truth and the optimal box with its corresponding better boundary score. Especially, our final regression loss consists of two parts: 
\begin{align}
    \label{loss: our-iou}
    \begin{aligned}
        L^{D\&R}_{IoU} = &\frac{1}{N_{pos}}\sum_{I}(\mathds{1}_{\{S'_{I}>S_{I}\}}L_{IoU}(B'_{I}, T_{I}) \\
        &+ \mathds{1}_{\{S_{I} \geqslant S'_{I}\}}L_{IoU}(B_{I}, T_{I})),
    \end{aligned}
\end{align}
where $\mathds{1}_{\{S_{I} \geqslant S'_{I}\}}$ is an indicator function, 
being $1$ if the original score is greater than the recombined one,
vice versa for $\mathds{1}_{\{S'_{I}>S_{I}\}}$.
The gradient of each boundary is selected to update network according to the higher IoU score between the original box and the recombined box.
Compared to the original IoU loss \Cref{loss: origin-iou} where gradients are back-propagated in local receptive fields,
\Cref{loss: our-iou} updates the network in context without extra parameterized computations.
As box in $B'_{I}$ is composed by boundaries from different boxes, 
features are updated in an instance-wise fashion. 
Note that there are no further parameters added in D\&R module. In short, we only change the way how gradient be updated.

\subsection{Semantic Consistency Module}
\label{subsec:reg-cls-consist}
Since the performance of our D\&R module to some extent depends on the box predictions of dense pixels inside an instance, an adaptive filtering method is required to help the network learning focus on positive pixels while rule out negative pixels. Namely, the labeling space of pixels inside an instance is expected to be consistent with their semantics.
Different from previous works \cite{OD-arXiv2019-FoveaBox,OD-ICCV2019-FCOS,OD-CVPR2019-Wang} which pre-define pixels around the center of the bounding box of an instance as the positive, our network evolves to learn the accurate labeling space without extra spatial assumptions in the training stage.

\begin{figure}[tb!]
    \centering
    \includegraphics[width=0.49\linewidth]{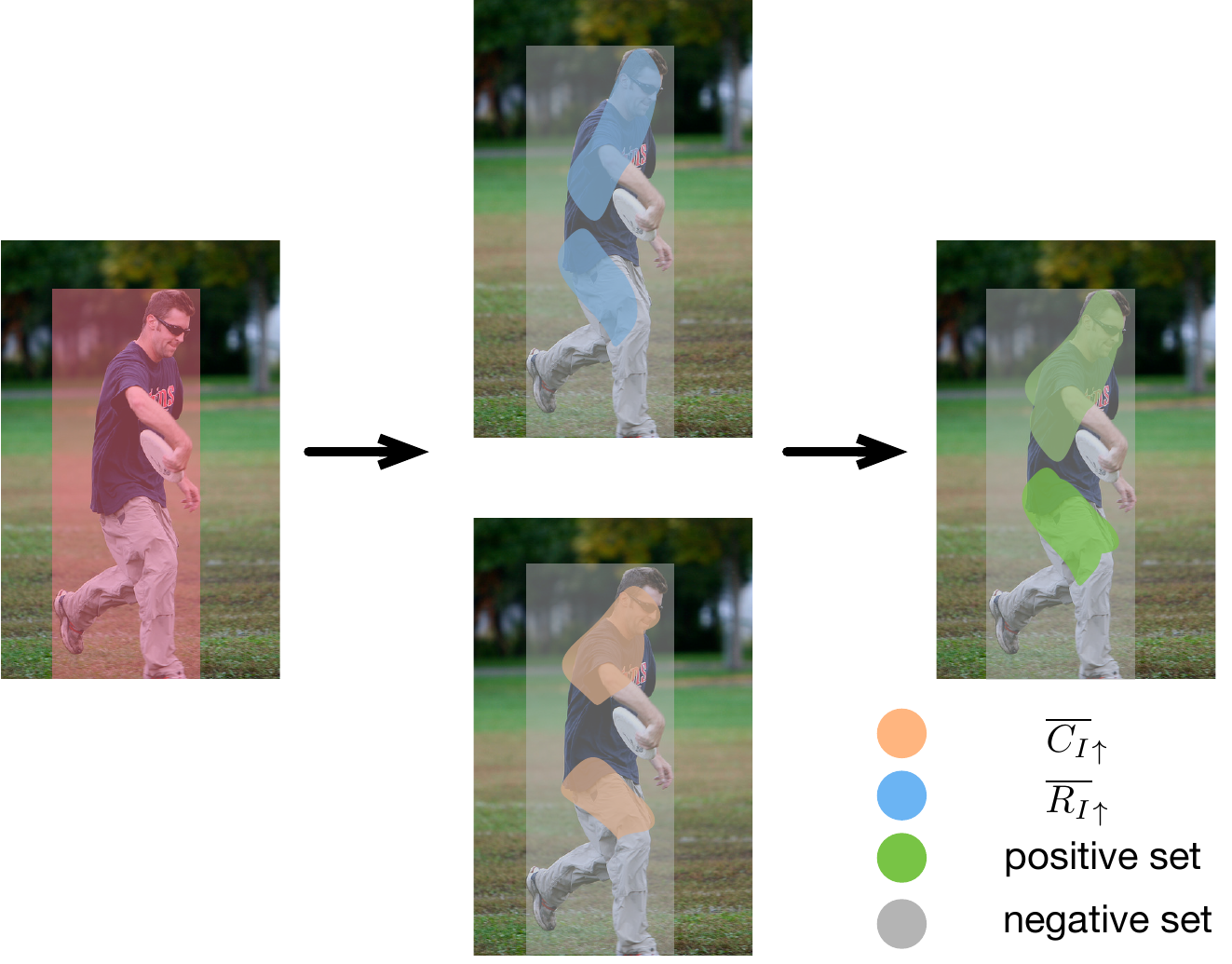} 
    \caption{\textbf{Visualized example of semantic consistency module.}
    The intersection regions of positive regression and positive classification sets are regarded as consistent targets.}  
    \label{figure: consistency-visualize}
\end{figure}

The formula of semantic consistency is expressed as:
\begin{align}
    \label{semanticeq}
    \begin{aligned}
        \left\{\begin{matrix}
            \overline{C_I}_{\downarrow} \bigcap  \overline{R_I}_{\downarrow}  \gets \text{negative}, 
            \\
            \\
            \overline{C_I}_{\uparrow} \bigcup  \overline{R_I}_{\uparrow}  \gets \text{positive}, \\
            \end{matrix}\right.\\
            c_{i} = \max_{j=0}^{g}(c_{j}) \in C_I,
    \end{aligned}
\end{align}      
where $R_I$ is the set of IoU scores between the ground-truth and the predicted boxes of pixels inside the instance $I$, 
$\overline{R_I}$ is the mean IoU score of the set $R_I$, $\overline{R_I}_{\downarrow}$ denotes pixels which have lower IoU confidence than the mean IoU $\overline{R_I}$. Inversely, $\overline{R_I}_{\uparrow}$ denotes pixels which have higher IoU confidence than $\overline{R_I}$.
The element $c_{i} \in C_I$ is the maximal classification score among all categories of the i-th pixel, and $g$ denotes the number of categories.
Similarly, $\overline{C_I}_{\downarrow}$ denotes pixels which have lower classification scores than the mean score of $C_{I}$.
Labels of categories are agnostic in this approach so that the predictions of incorrect categories will not be rejected during training.
Finally, as shown in \Cref{figure: consistency-visualize}, the intersection pixels in $\overline{R_I}_{\downarrow}$ and $\overline{C_I}_{\downarrow}$ are assigned negative, while the union pixels in $\overline{R_I}_{\uparrow}$ and $\overline{C_I}_{\uparrow}$ are assigned positive. Meanwhile, if pixels are covered by multiple instances, they prefer to represent the smallest instance.

More to the point, the filtering method determined by \Cref{semanticeq} is able to adaptively control the ratio of positive and negative pixels of instances with different sizes during the training stage, which have a significantly effect on the detection capability of the network. In the experiments, we investigate the performance of different fixed ratio, and then find that the adaptive selection by mean threshold performs best.

After the labels of pixels are determined autonomously according to the semantic consistency, the inner significance of each positive pixel is considered in the learning process of our network, similarly to the centerness score in FCOS\cite{OD-ICCV2019-FCOS}. Thus, our network is able to emphasize on more important part of an instance and is learnt more effectively. Especially, the inner significance of each pixel is defined as the IoU between the predicted box and the ground-truth. Then, 
an extra branch of estimating the semantic consistency of each pixel is added to the network supervised by the inner significance. 
The loss for semantic consistency is expressed as in \Cref{loss:cls-loss-consistency},
where $r_{i}$ is the output of semantic consistency branch. $IoU(p_{i}, p^*_{I})$ denotes the inner significance of each pixel.
\begin{align}
    \label{loss:cls-loss-consistency}
    L_{con} = \frac{1}{N_{pos}}\sum_{I}\sum_{i \in \overline{C_I}_{\uparrow} \bigcup  \overline{R_I}_{\uparrow}} CE(r_{i}, IoU(p_{i}, p^*_{I})).
\end{align}

Generally, the overall training loss is defined as:
\begin{align}
     \label{loss: loss-all}
     \begin{aligned}
         L = L_{cls} + L^{D\&R}_{reg} + L_{con},
     \end{aligned}
 \end{align}
where $L_{cls}$ is the focal loss as in \cite{OD-ICCV2017-RetinaNet}.

\section{Experiments}

\subsection{Experimental Setting}
\noindent\textbf{Dataset.} Our method is comprehensively evaluated on a challenging COCO detection benchmark\cite{lin2014microsoft}. 
Following the common practice of previous works\cite{OD-ICCV2019-FCOS,OD-ECCV2018-CornerNet,OD-ICCV2017-RetinaNet}, the COCO \textit{trainval35k} split (115K images) and the \textit{minival} split (5K images) are used for training  and validation respectively in our ablation studies. 
The overall performance of our detector is reported on the \textit{test-dev} split and is evaluated by the server. 
     
\noindent\textbf{Network Architecture.}
As shown in \Cref{figure: network-structure}, Feature Pyramid Network (FPN) \cite{OD-CVPR2017-FPN} is exploited as the fundamental detection network in our approach. The pyramid is constructed with the levels $P_{l}, l=3,4,...,7$  in this work. Note that each pyramid level has the same number of channels ($C$), where $C=256$. At the level $P_{l}$, the resolution of features is down-sampled by $2^{l}$ compared to the input size. 
Please refer to \cite{OD-CVPR2017-FPN} for more details. 
Note that four heads are attached to each layer of FPN. 
Apart from the regression and classification heads, a head for semantic consistency estimation is provided, consisting of a normal convolutional layer.
The regression targets of different layers are assigned in the same way as in \cite{OD-ICCV2019-FCOS}.
    
\noindent\textbf{Training Details.} Unless specified, all ablation studies take ResNet-50 as the backbone network. 
To be specific, the stochastic gradient descent (SGD) optimizer is applied and our network is trained for 12 epochs over 4 GPUs with a  minibatch of 16 images (4 images per GPU). 
Weight decay and momentum are set as 0.0001 and 0.9 respectively.
The learning rate starts at 0.01 and reduces by the factor of 10 at the epoch of 8 and 11 respectively.
Note that the ImageNet pre-trained model is applied for the network initialization.  
For newly added layers, we follow the same initialization method as in RetinaNet \cite{OD-ICCV2017-RetinaNet}. 
The input images are resized to the scale of $1333 \times 800$ as the common convention.
For comparison with state-of-the-art detectors, we follow the setting in \cite{OD-ICCV2019-FCOS} that the shorter side of images in the range from 640 to 800 are randomly scaled and the training epochs are doubled to 24 with the same reduction at epoch 16 and 22.
     
\noindent\textbf{Inference Details.} 
At post-processing stage, the input size of images are the same as the one in training. 
The predictions with classification scores $s>0.05$ are selected for evaluation.
With the same backbone settings, the inference speed of DDBNet is same as the detector in FCOS\cite{OD-ICCV2019-FCOS}.

            
  

\subsection{Overall Performance}

 \begin{table*}[tb]
        \centering
        \caption{\textbf{Comparison with state-of-the-art two stage and one stage Detectors}
        (\textit{single-model and single-scale results}).
        DDBNet outperforms the anchor-based detector \cite{OD-ICCV2017-RetinaNet} by 2.9\% AP with the same backbone.
        Compared with anchor-free models, DDBNet is in on-par with these state-of-the-art detectors.
        $^\dag$ means the NMS threshold is 0.6 and others are 0.5.}
        \label{tab: sota}
        \resizebox{1.02\linewidth}{!}{
            \begin{tabular}{l|l|lll|lll}
                \toprule    
                Method            & Backbone  & $AP$    & $AP_{50}$ & $AP_{75}$ & $AP_{S}$ & $AP_{M}$ & $AP_{L}$ \\
                \midrule
                \textbf{Two-stage methods:} &&&&&&\\
                Faster R-CNN w/ FPN       \cite{OD-CVPR2017-FPN}          & ResNet-101-FPN                                     & 36.2 & 59.1 & 39.0 & 18.2 & 39.0 & 48.2 \\
                Faster R-CNN w/ TDM       \cite{OD-arXiv2016-Shrivastava} & Inception-ResNet-v2-TDM \cite{szegedy2017inception} & 36.8 & 57.7 & 39.2 & 16.2 & 39.8 & 52.1 \\ 
                Faster R-CNN by G-RMI     \cite{OD-CVPR2017-Huang}        & Inception-ResNet-v2                                 & 34.7 & 55.5 & 36.7 & 13.5 & 38.1 & 52.0 \\
                RPDet                     \cite{OD-ICCV2019-RepPoints}    & ResNet-101-DCN                                      & 42.8 & 65.0 & 46.3 & 24.9 & 46.2 & 54.7 \\
                Cascade R-CNN             \cite{OD-CVPR2018-Cai}          & ResNet-101                                          & 42.8 & 62.1 & 46.3 & 23.7 & 45.5 & 55.2 \\
                \midrule \midrule
                \textbf{One-stage methods:} &&&&&&\\               
                YOLOv2                    \cite{OD-CVPR2017-YOLO9000}     & DarkNet-19\cite{OD-CVPR2017-YOLO9000}               & 21.6 & 44.0 & 19.2 & 5.0  & 22.4 & 35.5 \\
                SSD                       \cite{OD-ECCV2016-Liu}          & ResNet-101                                          & 31.2 & 50.4 & 33.3 & 10.2 & 34.5 & 49.8 \\
                DSSD                      \cite{OD-arXiv2017-DSSD}        & ResNet-101                                          & 33.2 & 53.3 & 35.2 & 13.0 & 35.4 & 51.1 \\
                FSAF                      \cite{OD-CVPR2019-Zhucc}        & ResNet-101                                          & 40.9 & 61.5 & 44.0 & 24.0 & 44.2 & 51.3 \\
                RetinaNet                 \cite{OD-ICCV2017-RetinaNet}    & ResNet-101-FPN                                      & 39.1 & 59.1 & 42.3 & 21.8 & 42.7 & 53.9 \\
                CornerNet                 \cite{OD-ECCV2018-CornerNet}    & Hourglass-104                                       & 40.5 & 56.5 & 43.1 & 19.4 & 42.7 & 53.9 \\
                ExtremeNet                \cite{OD-CVPR2019-Zhou}         & Hourglass-104                                       & 40.1 & 55.3 & 43.2 & 20.3 & 43.2 & 53.1 \\
                FCOS$^\dag$               \cite{OD-ICCV2019-FCOS}         & ResNet-101-FPN                                      & 41.5 & 60.7 & 45.0 & 24.4 & 44.8 & 51.6 \\
                FCOS$^\dag$               \cite{OD-ICCV2019-FCOS}         & ResNeXt-64x4d-101-FPN                               & 43.2 & 62.8 & 46.6 & 26.5 & 46.2 & 53.3 \\
                FCOS$^\dag$ w/improvements\cite{OD-ICCV2019-FCOS}         & ResNeXt-64x4d-101-FPN                               & 44.7 & 64.1 & 48.4 & 27.6 & 47.5 & 55.6 \\
                \midrule
                DDBNet (Ours)                                         & ResNet-101-FPN          & 42.0 & 61.0 & 45.1 & 24.2 & 45.0 & 53.3 \\
                DDBNet (Ours)                                         & ResNeXt-64x4d-101-FPN   & 43.9 & 63.1 & 46.7 & 26.3 & 46.5 & 55.1 \\
                DDBNet (Ours)$^\mathsection$                          & ResNeXt-64x4d-101-FPN   & \textbf{45.5} & \textbf{64.5} & \textbf{48.5} & \textbf{27.8} & \textbf{47.7} & \textbf{57.1} \\
                \bottomrule
            \end{tabular}
        }
        \scriptsize{
            \begin{flushleft}
                $^\mathsection$ GIoU\cite{Rezatofighi_2018_CVPR} and Normalization methods of `improvements' proposed in FCOS\cite{OD-ICCV2019-FCOS} are applied, ctr.sampling in `improvements'\cite{OD-ICCV2019-FCOS} are not compatible with our setting and we do not use.
            \end{flushleft}
        }
        \vspace{-.4in}
    \end{table*}

We compare our model denoted as DDBNet with other state-of-the-art object detectors on the \textit{test-dev} split of COCO benchmark, as listed in \Cref{tab: sota}.
Compared to the anchor-based detectors, our DDBNet shows its competitive detection capabilities. Especially, it outperforms RetinaNet\cite{OD-ICCV2017-RetinaNet} by 2.9\% AP. When it comes to the anchor-free detectors, especially detectors such as FCOS\cite{OD-ICCV2019-FCOS} and CornerNet\cite{OD-ECCV2018-CornerNet} benefiting from the point-based representations, our DDBNet achieves performances gains of 0.5\% AP and 1.5\% AP respectively.
Based on the ResNeXt-64x4d-101-FPN backbone \cite{SPEED-CVPR2017-ResNeXt},
DDBNet works better than \cite{OD-ICCV2019-FCOS} with a 0.7\% AP gain.
Especially for large objects, our DDBNet gets 55.1\% AP, better than 53.3 \% reported in FCOS\cite{OD-ICCV2019-FCOS}.
We also apply part of 'improvement' methods proposed in FCOS to DDBNet and gets 0.8\% better than the FCOS with all `improvements' applied.
To sum up, compared to detectors exploiting point-based representations, our DDBNet can similarly benefit from the mid-level boundary representations without heavy computation burdens. Furthermore, DDBNet is compared to several two stage models. It overpasses \cite{OD-CVPR2017-FPN} by a large margin.

\vspace{-.1in}
\subsection{Ablation Study}

In this section, we explore the effectiveness of our
method, including two main modules of box D\&R module and semantic consistency module. Additionally, we conduct in-depth analysis of the performance metrics of our method.

\subsubsection{Comparison with Baseline Detector}\mbox{}

It should be noted that FCOS detector\cite{OD-ICCV2019-FCOS} without the centerness branch in both training and inference stages is taken as our baseline. 
Here we conduct in-depth analysis of the performance metrics of our method. 

\noindent\textbf{Box D\&R module.} 
As shown in \Cref{tab:b&r-ablation}, by incorporating the D\&R module into the baseline detector, a 1.2\% $AP$ gain is obtained, which proves that our D\&R module can boost the overall performance of the detector. Especially for the $AP_{75}$, a 1.4\% improvement is achieved, which means that D\&R performs better on localization even in a strict IOU threshold. Furthermore, D\&R module achieves a better performance on large instances according to the large gain on $AP_L$. 
With explicit boundary analysis, large instances are often surrounded by numbers of predicted boxes. As a result, it gets easier to find the well-aligned boundaries, then the boxes re-organization can be more effective.  
Compared to the baseline results in metrics including $AP_{50}$, $AP_{S}$ and $AP_{M}$, D\&R obtains stable performance gains respectively, which shows the stability of our proposed module.
By breaking the atomic boxes into boundaries, D\&R module makes each boundary find the better optimization direction. The optimization of boundary is not limited by the box its in, instead of depending on a sorted of related boxes. Generally, by adjusting the boundary optimization, the detection network is learnt better.  
    \begin{table*}[t]
        \centering
        \caption{\textbf{Ablative experiments for DDBNet on the COCO \textit{minival} split.} We evaluate the improvements brought by the Box Decomposition and Recombination(D\&R) module and the semantic consistency module.}
        \resizebox{.68\linewidth}{!}{
            \begin{tabular}{ccc|l|ll|lll}
                \toprule    
                \multicolumn{3}{c|}{ Modules }    & $AP$    & $AP_{50}$ & $AP_{75}$ & $AP_{S}$ & $AP_{M}$ & $AP_{L}$ \\
                Baseline & D\&R & Consistency     &  & & & & & \\
                \midrule
                \checkmark &            &            & 33.6          & 53.1          & 35.0          & 18.9          & 38.2          & 43.7  \\ 
                \checkmark & \checkmark &            & 34.8          & 54.0          & 36.4          & 19.7          & 39.0          & 44.9  \\
                \checkmark &            & \checkmark & 37.2          & 55.4          & 39.5          & 21.0          & 41.7          & 48.6  \\ 
                \checkmark & \checkmark & \checkmark & \textbf{38.0} & \textbf{56.5} & \textbf{40.8} & \textbf{21.6} & \textbf{42.4} & \textbf{50.4}  \\
                \bottomrule
            \end{tabular}
        }
        \label{tab:b&r-ablation}
    \end{table*}

\noindent\textbf{Semantic Consistency module.}
The semantic consistent module described in \Cref{subsec:reg-cls-consist} presents an adaptive filtering method. It forces our detection network into autonomously focusing on positive pixels whose semantics are consistent with the target instance. As shown in \Cref{tab:b&r-ablation}, the semantic consistency module contributes to a significant performance gain of 3.6\% $AP$ compared to the baseline detector. This variant surpasses the baseline by large margins in all metrics. 
Due to that the coarse bounding boxes would contain backgrounds and distractors inevitably, the network is learnt with less confusion about the targets when equips our adaptive filtering module. More ablation analysis on semantic consistency module is provided in \Cref{subsubsec:ana-sc}.
        
\noindent\textbf{Cooperation makes better.}
In our final model denoted as DDBNet, the semantic consistency module first filters out a labeling space of pixels inside each instance that is strongly relative to the geometric and semantic characteristics of the instance. The box predictions of the filtered positive pixels are further optimized by the D\&R module, leading to more accurate detection results. Consequently, DDBNet achieves 38\% AP, better than all the variants in \Cref{tab:b&r-ablation}. Our method boosts detection performance over the baseline by 2.7\%, 4.2\%, and 6.7\% respectively on $AP_S$, $AP_M$, $AP_L$.

\subsubsection{Analysis on D\&R Module.}\mbox{}

\begin{figure}[tb!]
    \centering
    \pgfplotsset{
        width =0.58\linewidth,
        height=0.36\linewidth
    }
\begin{tikzpicture}[scale=1]
\begin{axis}[minor tick num=0,
xmin=-0.5, xmax=12,
ymin=0.25, ymax=0.95,
mark size=1.0pt,
ytick={0.3,0.5,...,0.9},
xlabel={epoch},
ylabel={IoU},
xlabel near ticks,
ylabel near ticks,
legend style={
  draw=none,
  at={(1.0,.28)},
  anchor=west,
  legend columns=1},
]
\addplot [color=red, only marks, mark=o, line width=1.0pt]
    plot [error bars/.cd, y dir = both, y explicit]
    table [x={iter},  y={dr-mean}, y error={var}]  {dr-iou.mat};
\addplot [color=blue, only marks, mark=o, line width=1.0pt]
    plot [error bars/.cd, y dir = both, y explicit]
    table [x={iter},  y={iou-mean}, y error={var}]     {origin-iou.mat};
\legend{IoU w/~D\&R, IoU w/o~D\&R}
\end{axis}
\end{tikzpicture}
\caption{\textbf{Average IoU scores for all predicted boxes during the training.}
    The red points denote the IoU scores with D\&R module while the blue points are the IoU scores without optimization.
    Vertical lines indicate the variance of IoU scores.}
    \label{figure: mean-iou-compare}

\end{figure}
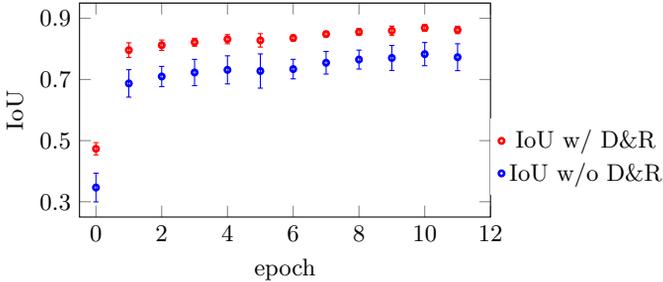

\noindent\textbf{Statistical comparison with conventional IoU Loss.}
As we mentioned in \Cref{subsec:box-decom-recom}, IoU loss with D\&R updates the gradient according to the optimal boundary scores.
To confirm the stability of D\&R module, we plot the average IoU scores and variances of boxes before and after D\&R respectively. 
We can see that with D\&R module, the average values of IoU scores are higher than the means of origin IoU scores by a large margin around 10\% in the whole training schedule, as in \Cref{figure: mean-iou-compare}. 
At the start of training, the mean of optimal boxes gets 0.47  which is better than 0.34 of origin boxes. 
As training goes on, both average scores of origin and optimal boxes increase and remain at 0.77 and 0.86 at the end.
Variances of IoU scores with D\&R are much lower than the origin IoU scores, which indicates D\&R module improves the overall quality of boxes and provides better guidance for training.

\begin{figure*}[t!]
    \centering
    \includegraphics[width=0.88\linewidth]{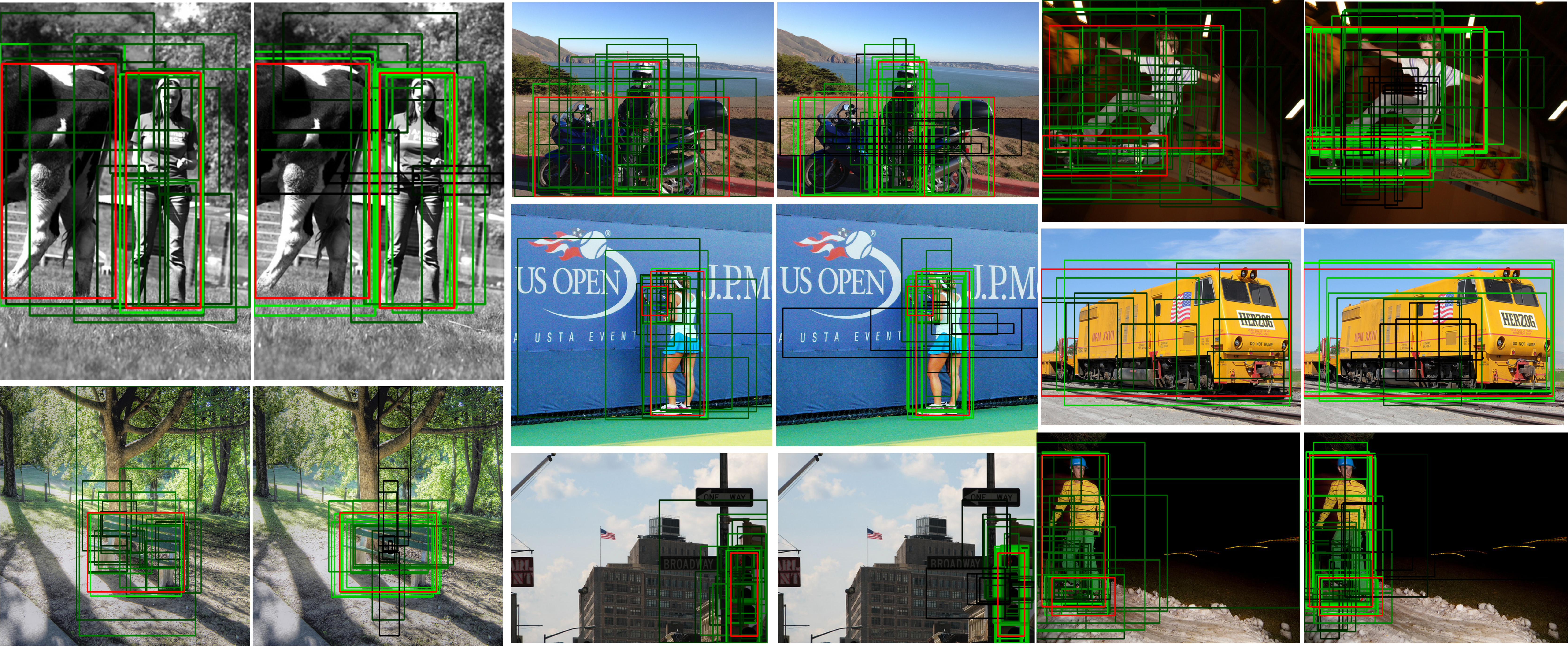} 
    \caption{
        \textbf{Illustration of improved box predictions provided by our DDBNet.}
        We visualize the boxes before the decomposition (left images of the pairs) and the boxes after the recombination (right images of the pairs). \textbf{Red}: ground-truth boxes. \textbf{Green}: the predictions, where the lighter colors indicate higher IoU scores. \textbf{Black}: the boxes with low score, which will be masked according to the regression loss.
        Boxes ranked by D\&R module are much better organized than the origin boxes and the localizations are much correlated to the instances.
        All the results are from DDBNet with ResNet-50 as backbone on $trainval35k$ split.}
    \label{figure: dr-appendix-visualize}
\end{figure*} 

\noindent\textbf{Visualization on D\&R module.} 
We provide some qualitative results of box predictions before and after incorporating the D\&R module into the baseline detector, 
as shown in \Cref{figure: dr-appendix-visualize}.
For clear visualization, we plot origin boxes and boxes after recombination individually. Predictions are presented in green and the lighter colors indicate higher IoU scores.
With D\&R module, boundaries are recombined together to obtain a tighter box of each instance.
The distribution of boxes after D\&R module is fitter than the origin boxes which is robust than the conventional regression.
As we mentioned in \Cref{subsec:box-decom-recom}, there exists recombined low-rank boxes with boundary scores lower than the origin.
These boundaries are masked according to the \Cref{loss: our-iou}.

\label{subsubsec:ana-sc}
\subsubsection{Analysis on Semantic Consistency}\mbox{}

            \begin{table}[tb!]
                \centering
                \caption{\textbf{Comparison among different positive assignment strategies.} 
                `None' means no sampling method is applied.
                `PN' denotes as the definition in \cite{OD-arXiv2019-FoveaBox}, 
                 which means center regions are positive and others are negative. 
                `PNI' is the sampling used in \cite{OD-CVPR2019-Zhucc,OD-CVPR2019-Wang},
                ignore regions are added between positive and negative.
                Note that the consistency term is not included in this table.}
                \begin{tabular}{c|l|ll|lll}
                    \toprule    
                    Settings   & $AP$    & $AP_{50}$ & $AP_{75}$ & $AP_{S}$ & $AP_{M}$ & $AP_{L}$ \\
                    \midrule
                    None        & 33.6     & 53.1    & 35.0    & 18.9     & 38.2    & 43.7  \\ 
                    PN        & 34.2     & 53.2    & 36.3    & 20.8     & 38.9    & 44.2  \\
                    PNI       & 33.7     & 53.0    & 35.5    & 17.9     & 38.3    & 44.1  \\
                    Ours      & \textbf{35.3}     & \textbf{55.4}    & \textbf{37.1}    & \textbf{20.9}     & \textbf{39.6}    & \textbf{45.9}  \\ 
                    \bottomrule
                \end{tabular}
                \label{tab:consistency-ablation}
            \end{table}
\noindent\textbf{Dynamic or predefined positive assignment.} 
To further show the superiority of dynamic positive assignment in semantic consistency module, we investigate other variants using different predefined strategies mentioned in previous works. FoveaBox\cite{OD-arXiv2019-FoveaBox} (denoted as `PN') applies center sampling in their experiments to improve the detection performance. 
This center sampling method defines the central area of a target box based on a constant ratio as positive while the others as negative.
`PNI' is taken used in \cite{OD-CVPR2019-Zhucc, OD-CVPR2019-Wang} which exploits positive, ignore and negative regions for supervised network training.
According to the result in \Cref{tab:consistency-ablation},
`PN' (second line) gets slight improvement compared to the baseline where no sampling method is adopted. 
So restricting the searching space to the central area makes sense in certain cases and indeed helps improve object detection.
But the 'PNI' gets a lower performance, especially on $AP_S$. Namely, adding an ignore region between the ring of negative areas and the central positive areas does not further improve the performance and gets a large drop on the detection of small objects.
The limited number of candidates of small objects and the lower ratio of positive candidates in `PNI' result in the poor detection capability.
Contrastively, our proposed filtering method does not need to pre-define the spatial constraint while show best performances in all metrics.

\begin{table}[tb!]
    \centering
    \caption{\textbf{Comparison among different ratio settings.} 
    where $c$ is the sampling ration for each instance.}
    \begin{tabular}{c|l|ll|lll}
        \toprule    
        ratios   & $AP$    & $AP_{50}$ & $AP_{75}$ & $AP_{S}$ & $AP_{M}$ & $AP_{L}$ \\
        \midrule
        $c=0.4$                    & 34.6 & 54.2 & 36.6 & 19.1 & 38.5 & 45.2  \\
        $c=0.5$                    & 34.1 & 53.5 & 35.9 & 19.2 & 38.4 & 44.2  \\ 
        $c=0.6$                    & 34.7 & 54.2 & 36.5 & 19.0 & 38.7 & 45.5  \\ 
        $c=0.7$                    & 35.1 & 54.6 & \textbf{37.1} & 19.3 & 39.1 & 45.7  \\ 
        $mean$                     & \textbf{35.3}     & \textbf{55.4}    & \textbf{37.1}    & \textbf{20.9}     & \textbf{39.6}    & \textbf{45.9}  \\ 
        \bottomrule
    \end{tabular}
    \label{tab:consistency-ratio}
\end{table}
\noindent\textbf{Adaptive or constant ratio.} 
As mentioned in \cref{subsec:reg-cls-consist}, we investigate the constant ratio to replace the adaptive selection by mean. Four variants are obtained where the constant ratio is set from 0.4 to 0.7. 
For instance $I$ with $M$ candidates, top $\left \lfloor c \times M \right \rfloor$  candidates are considered as positive, and others are negative, where $c$ is the constant sampling ratio applied to all instances. 
As shown in \Cref{tab:consistency-ratio}, these results indicate that the adaptive way in our method is better than the fixed way to select positives from candidates.


\section{Conclusion}
  We propose an anchor-free detector DDBNet, which firstly proposes
  the concept of breaking boxes into boundaries for detection.
  The box decomposition and recombination optimizes the model training by uniting atomic pixels and updating in a bottom-up manner.
  We also re-evaluate the semantic inconsistency during training, 
  and provide an adaptive perspective to solve this problem universally with no predefined assumption. Finally, DDBNet achieves a state-of-the-art performance with inappreciable computation overhead for object detection.

\newpage
\appendix
This is our supplementary material which includes more experiments on semantic consistency and performance analysis to show the effectiveness of our work.

    
\section{Analysis on Semantic Consistency Module}
\begin{figure*}
            \centering
            \includegraphics[width=0.85\linewidth]{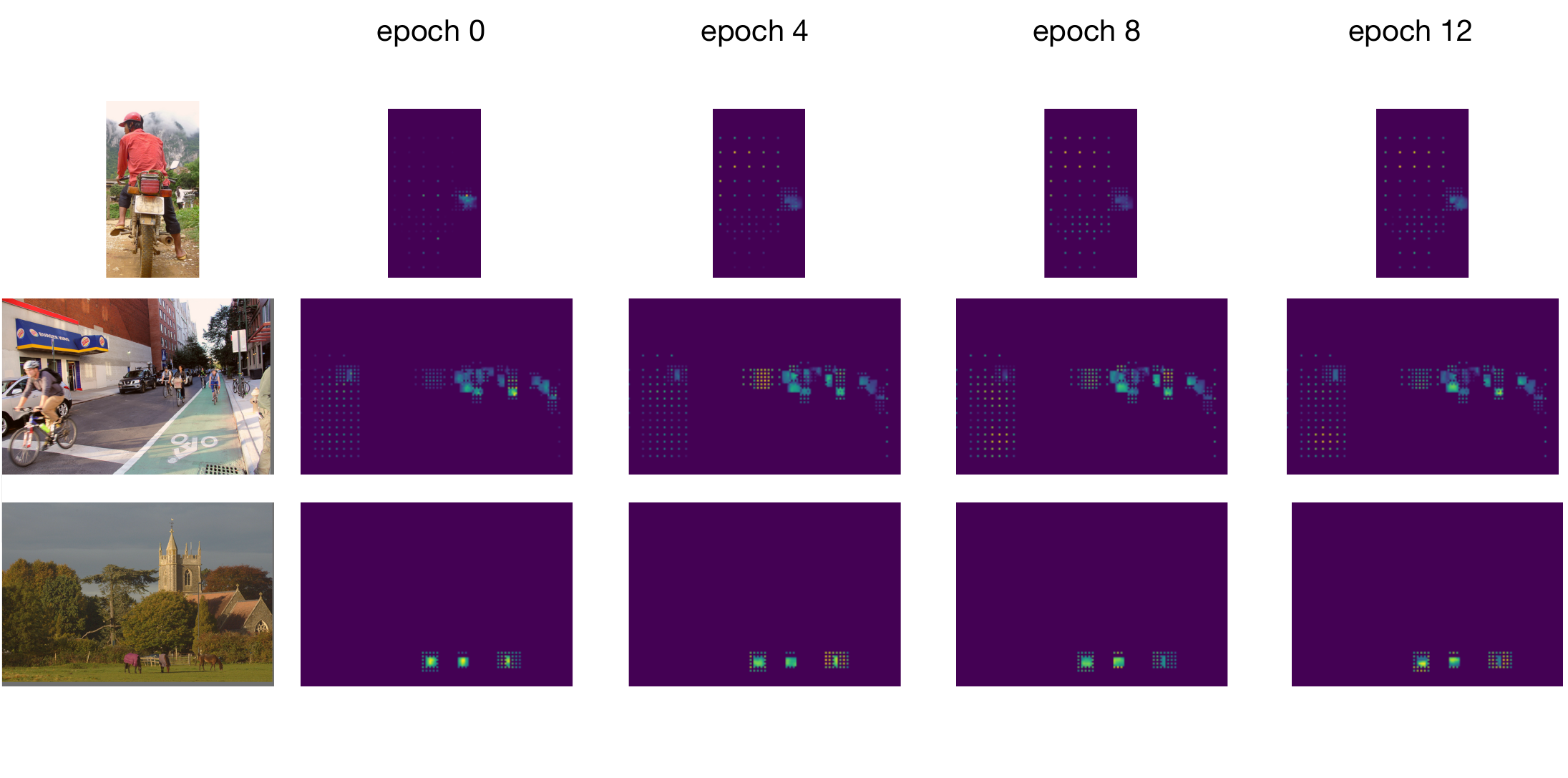} 
            \caption{Visualized examples of semantic  consistency module.
            The left image of each row is the training data select from COCO $trainval35k$.
            Rest images on the right are the heatmaps of sampled points with semantic consistency module at different training epoch.
            Note that, entries of heatmaps represent the product results of IoU scores and classification scores.
            Sampled points with high IoU scores and high classification scores are highlighted in the heatmaps.
            Sampled points with low IoU scores or low classification scores are in dark colors.\textit{Better viewed in colors and zoom in.}
            }
            \label{figure: consistency-appendix-visualize}
\end{figure*}

\begin{figure*}
            \centering
            \includegraphics[width=0.85\linewidth]{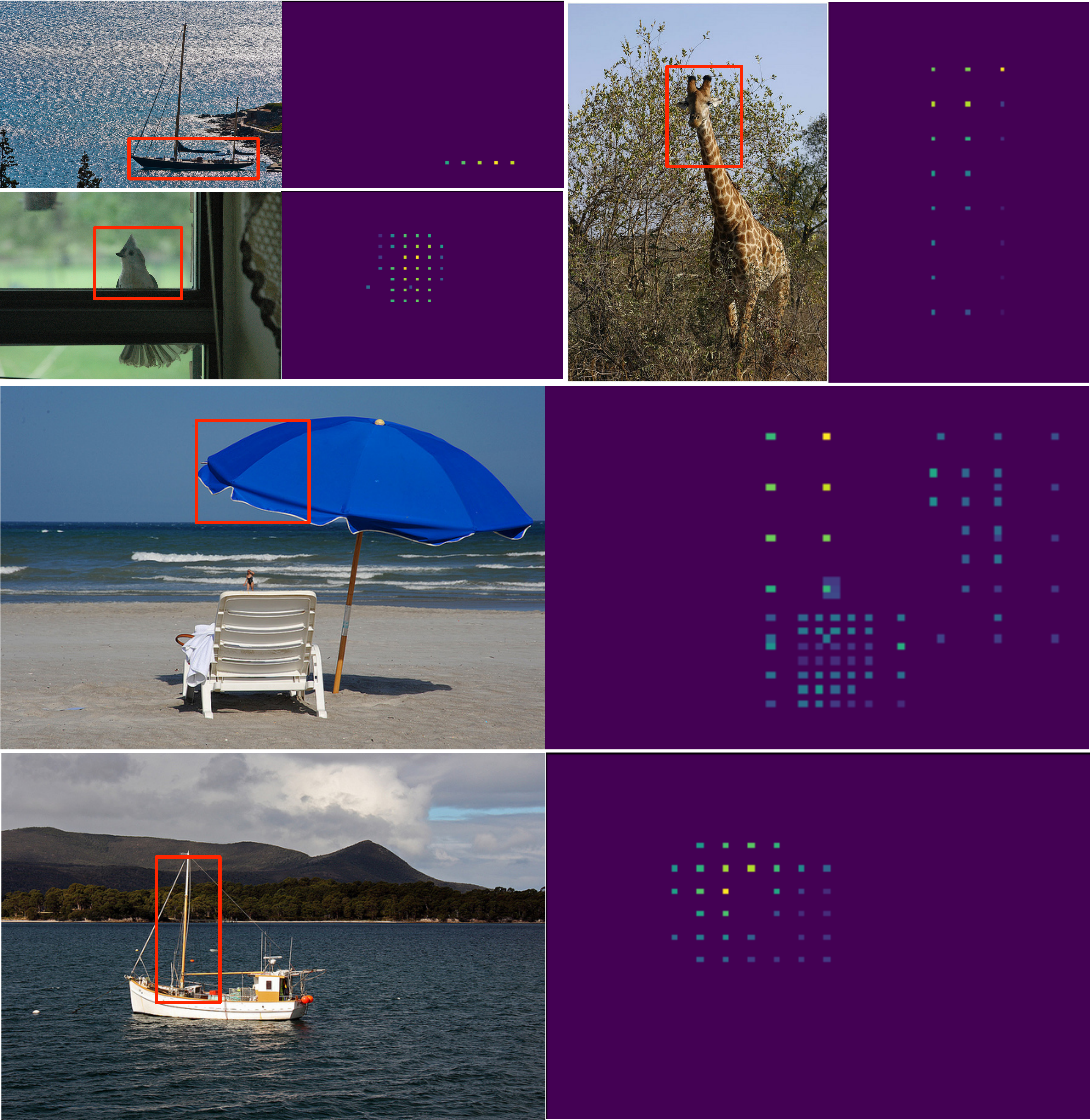} 
            \caption{Visualization on center inconsistency examples. 
            Sample points with high IoU scores and high classification scores are highlighted. 
            The corresponding areas preferred by the semantic consistency module are marked as red boxes on the images.
            Images are select from COCO $trainval35k$ and evaluated with the trained model with ResNet-50 as the backbone.
            }
            \label{figure: sc-visualize}
\end{figure*}

We visualize the dynamic consistency at different epoches to see how semantic consistency affects on the learning targets.
As shown in \Cref{figure: consistency-appendix-visualize}, 
the sampled points in each epoch with both high classification scores among categories and high IoU scores are highlighted, 
named high consistency samples.
The low consistency samples are appear in dark colors. 
Part of sample points at initial stage is not locate at the instance, as the model is not robust at the beginning.
With the semantic consistency module, the learned positive samples are progressively distributed at the semantic area of the instance.
As the training going on, 
high consistency samples become robust and appear in lighter colors.
We also evaluate to see how inconsistency problem be solved by our method. 
Some qualitative results are presented in \cref{figure: sc-visualize},
the typical inconsistency in which center-like annotations cannot handle presented in \Cref{figure:inconsistency-example} are improved to a large extent. 
By utilizing the segmentation annotations, 
we found that the proportion of samples locate on background reduced around 15\% (from 51.7\% at initial to 36.1\% when training finished) with the semantic consistency module. 

\section{Precision-Recall curves}
\begin{figure}[ht]
            \centering
            \subfloat[]{\label{figure: test}{\includegraphics[width=0.5\textwidth]{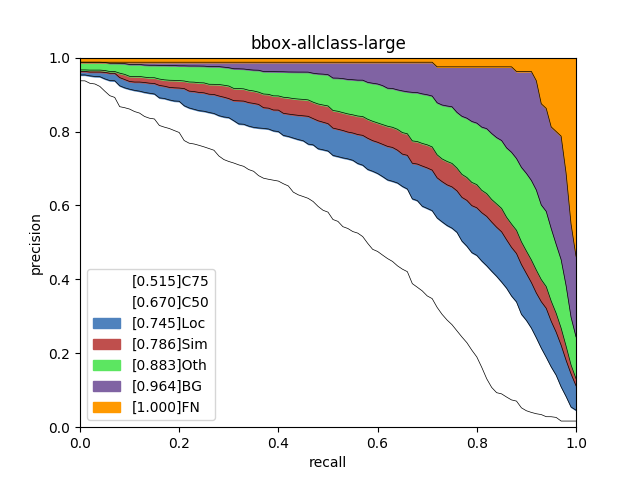}}}
            \subfloat[]{\label{figure: test}{\includegraphics[width=0.5\textwidth]{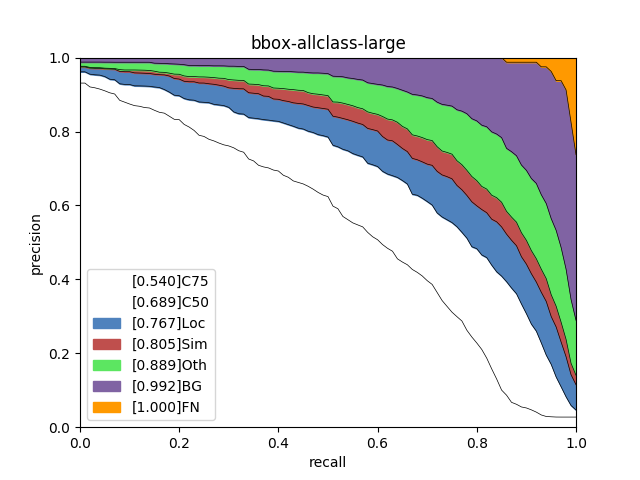}}}
            \vspace{-1cm}
            \subfloat[]{\label{figure: test}{\includegraphics[width=0.5\textwidth]{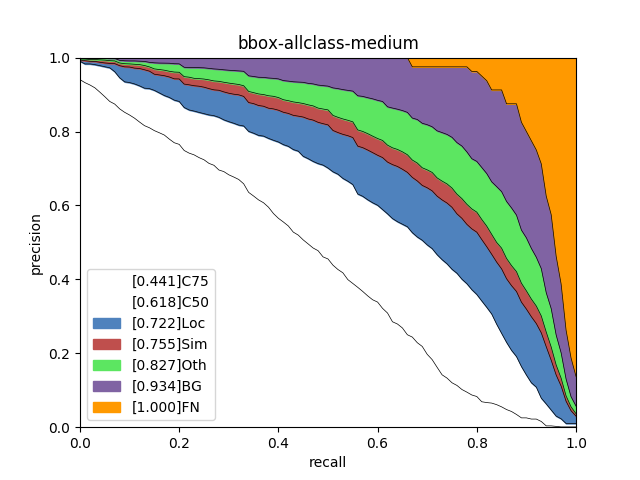}}}
            \subfloat[]{\label{figure: test}{\includegraphics[width=0.5\textwidth]{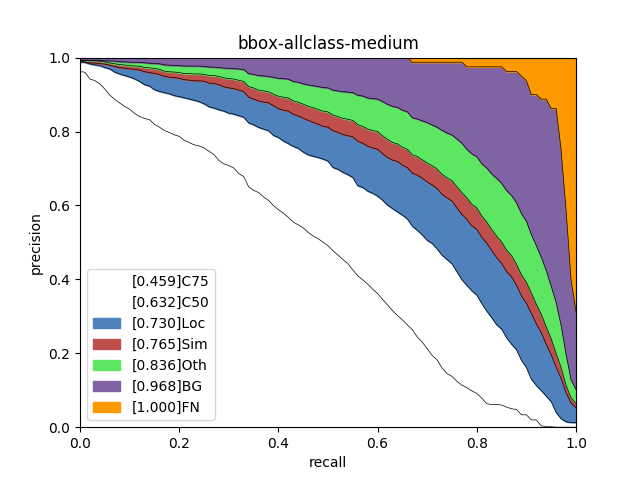}}}
            \vspace{-1cm}
            \subfloat[]{\label{figure: test}{\includegraphics[width=0.5\textwidth]{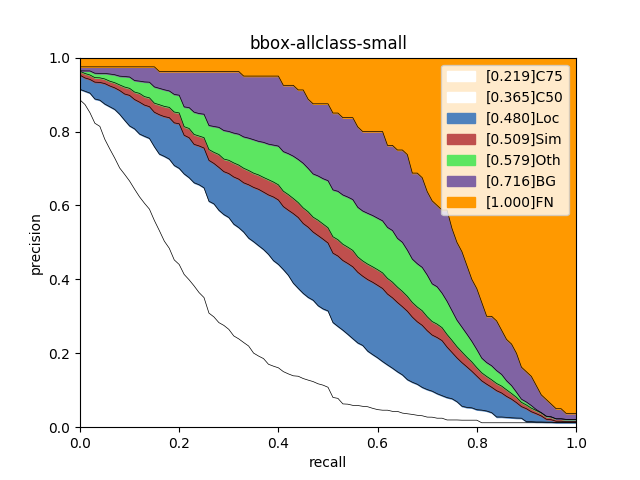}}}
            \subfloat[]{\label{figure: test}{\includegraphics[width=0.5\textwidth]{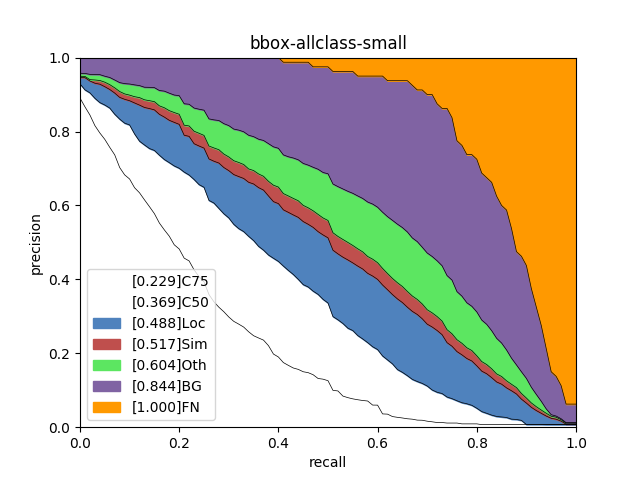}}}
            \caption{\textbf{Precision Recall Curves.}
            Precision-recall(PR) curves of FCOS\cite{OD-ICCV2019-FCOS} and DDBNet under different evaluation settings provided by \cite{SEGM-ECCV2014-COCO} on the \textit{minival} split with ResNet-50 as backbone. (a)(c)(e): Evaluation results in FCOS. (b)(d)(f): Evaluation results in DDBNet. DDBNet gets better performance under the strict evaluation settings. Especially, we find out that DDBNet works much robust after all background and class confusions removed.}
            \label{figure:pr-curve}
\end{figure}

The precision-recall(PR) curves of FCOS\cite{OD-ICCV2019-FCOS} and DDBNet under different evaluation settings provided by \cite{SEGM-ECCV2014-COCO} on the \textit{minival} split are shown in \Cref{figure:pr-curve}.
PR curves were plotted for small-, medium- and large-scale objects in two models. The area in orange indicates the false negative(FN) portion of the evaluated dataset, which can be considered as the PR with all errors removed. The purple area presents the falsely detected objects. 
We can see that the area of orange in DDBNet is much lower than the one in FCOS\cite{OD-ICCV2019-FCOS}, which means DDBNet is much robust after all background and class confusions removed.

%

\clearpage
%
%
\bibliographystyle{splncs04}
\bibliography{ref/Top,ref/LEARN,ref/DL,ref/CV,ref/SEGM,ref/Signal,ref/SPEED,ref/addition}
\end{document}